\documentclass[sigconf]{acmart}

\setcopyright{none}

\setcopyright{acmcopyright}
\acmConference[CF '21]{Computing Frontiers Conference}{May 11--13, 2021}{Virtual Conference, Italy}
\acmBooktitle{Computing Frontiers Conference (CF '21), May 11--13, 2021, Virtual Conference, Italy}
\acmPrice{15.00}
\acmDOI{10.1145/3457388.3458656}
\acmISBN{978-1-4503-8404-9/21/05}

\usepackage[ruled,vlined]{algorithm2e}
\usepackage{comment}
\usepackage{subcaption}
\usepackage{multirow}
\usepackage{url}
\usepackage{relsize}
\usepackage{tabularx}

\begin{document}

\title{Ultra-compact Binary Neural Networks for Human Activity Recognition on RISC-V Processors}

\author{Francesco Daghero}
\email{francesco.daghero@polito.it}
\affiliation{
   \institution{Department of Control and Computer Engineering, Politecnico di Torino, Turin, Italy }
}

\author{Chen Xie}
\email{chen.xie@polito.it}
\affiliation{
   \institution{Department of Control and Computer Engineering, Politecnico di Torino, Turin, Italy }
}

\author{Daniele Jahier Pagliari}
\email{daniele.jahier@polito.it}
\affiliation{
   \institution{Department of Control and Computer Engineering, Politecnico di Torino, Turin, Italy }
}

\author{Alessio Burrello}
\email{alessio.burrello@unibo.it}
\affiliation{
   \institution{Department of Electrical, Electronic and Information Engineering, University of Bologna, Bologna, Italy}
}

\author{Marco Castellano}
\email{marco.castellano@st.com}
\affiliation{
   \institution{STMicroelectronics, Cornaredo, Italy}
}

\author{Luca Gandolfi}
\email{luca.gandolfi1@st.com}
\affiliation{
   \institution{STMicroelectronics, Cornaredo, Italy}
}

\author{Andrea Calimera}
\email{andrea.calimera@polito.it}
\affiliation{
   \institution{Department of Control and Computer Engineering, Politecnico di Torino, Turin, Italy }
}

\author{Enrico Macii}
\email{enrico.macii@polito.it}
\affiliation{
   \institution{Department of Control and Computer Engineering, Politecnico di Torino, Turin, Italy }
}

\author{Massimo Poncino}
\email{massimo.poncino@polito.it}
\affiliation{
   \institution{Department of Control and Computer Engineering, Politecnico di Torino, Turin, Italy }
}

\renewcommand{\shortauthors}{F. Daghero et al.}

\begin{abstract}
Human Activity Recognition (HAR) is a relevant inference task in many mobile applications. State-of-the-art HAR at the edge is typically achieved with lightweight machine learning models such as decision trees and Random Forests (RFs), whereas deep learning is less common due to its high computational complexity. In this work, we propose a novel implementation of HAR based on deep neural networks, and precisely on Binary Neural Networks (BNNs), targeting low-power general purpose processors with a RISC-V instruction set. BNNs yield very small memory footprints and low inference complexity, thanks to the replacement of arithmetic operations with bit-wise ones. However, existing BNN implementations on general purpose processors impose constraints tailored to complex computer vision tasks, which result in over-parametrized models for simpler problems like HAR. Therefore, we also introduce a new BNN inference library, which targets \textit{ultra-compact} models explicitly. With experiments on a single-core RISC-V processor, we show that BNNs trained on two HAR datasets obtain higher classification accuracy compared to a state-of-the-art baseline based on RFs. Furthermore, our BNN reaches the same accuracy of a RF with either less memory (up to 91\%) or more energy-efficiency (up to 70\%), depending on the complexity of the features extracted by the RF.
\end{abstract}



\maketitle

\section{Introduction}
Machine Learning (ML) plays an increasingly important role in many Internet of Things (IoT) applications, ranging from computer vision to time-series processing~\cite{sze2017efficient,chen2019deep,daghero2020energy,shi2016edge}. Edge computing, as a paradigm to host data-analytics as close as possible to end devices, may offer several advantages compared to the standard cloud-centric approach. In fact, by not relying completely on the network, it reduces the issues of
long or unpredictable latency in presence of slow or unstable connectivity. Moreover, it is also potentially beneficial for energy and privacy, as it avoids the power-hungry transmission of large chunks of (possibly sensitive)  user data through wireless channels~\cite{sze2017efficient,chen2019deep,shi2016edge}.
The main obstacle to the widespread adoption of edge computing in ML is that moving inference tasks on tiny, low-cost, and memory-/energy-constrained processors  requires ultra-compact implementations of complex ML algorithms. 

Human Activity Recognition (HAR) based on Inertial Measurement Units (IMUs) is a popular ML-based application that benefits from edge computing, being commonly found in devices such as smartwatches. Most current HAR solutions at the edge are based on classic ML, and in particular on simple and HW-friendly tree-based algorithms, such as decision trees and Random Forests (RFs)~\cite{stsensor,fan2013human,balli2019human}.
Deep learning (DL) methods are less common, mostly because their high memory and computation requirements do contrast with the limited resources available on low-power edge devices~\cite{sze2017efficient,crime}.
 
Binary Neural Networks (BNN) could overcome this limitation, thanks to their very limited memory requirements and to the usage of low-cost binary operations. However, in literature, BNNs have never been applied to HAR, and in general, they have been mostly implemented with custom HW accelerators~\cite{xnorengine,bnn_accel}. Implementations on general purpose processors are rare, and the few existing impose important restrictions on the minimum supported size of the network~\cite{pulpNN,bmxnet,zhang2019dabnn}.
Specifically, the number of channels in each layer is constrained to be a multiple of 32,
in order to simplify memory transfers and binary operations on 32-bit registers commonly found in edge processors. While these restrictions are suitable for complex computer vision applications, we show that on simpler tasks like HAR they make the resulting BNNs more complex than what is actually needed to reach high accuracy.

In this work, we describe a new library for BNN inference that targets RISC-V processors, and is specifically targeted at what we call \textit{ultra-compact networks}, i.e., with fewer than 32 channels per layer.
To the best of our knowledge, our work introduces both the first application of BNNs to HAR and the first library for general purpose HW focusing on small BNNs.
With results on two different HAR datasets (one public and one proprietary) and using the 32-bit single-core RISC-V Quentin~\cite{quentin}, we show that our library is able to reduce the number of parameters by 7\% and the number of cycles by 21\% for the same accuracy, compared to a standard BNN library limited to multiples of 32 channels.
Moreover, our BNN implementation is Pareto-optimal compared to a solution based on RFs, outperforming the latter either in terms of inference latency (and hence energy consumption) or in terms of memory occupation, depending on the dataset and on the feature set used by the RF. Specifically, we are able to reduce the latency/energy by up to 70\% on one dataset, and the memory occupation by up to 91\% on the other compared to the RF approach, with no accuracy loss.

\section{Background and Related Works}

\subsection{Human Activity Recognition}
HAR is typically formulated as a classification task, aimed at identifying human activities (e.g., standing, laying, walking, running, etc.) based on inertial sensor measurements~\cite{anguita2012human}. 
In recent years, research on ML techniques for HAR has become very active and relevant on a variety of fields, such as health monitoring and fall detection.  

One of the first HAR approaches targeting edge devices explicitly is proposed in~\cite{anguita2012human}, where the authors implement a Support Vector Machine (SVM) classifier with fixed-point arithmetic, in order to reduce the computational cost with minimal accuracy loss, with respect to a floating point version. More recently, \cite{micucci2017unimib} presented a novel HAR dataset composed of acceleration signals gathered from smartphones, and then benchmarked this dataset with four classifiers: k-Nearest Neighbour (KNN), SVM, fully-connected Artificial Neural Networks (ANN) and Random Forests (RF). Experiments were performed on four different classification tasks (in terms of number of classes to recognize), and
each algorithm was evaluated either on raw data and on magnitude-based features. Results showed that, in tasks distinguishing among different types of Activities of Daily Living (ADLs), RFs obtain the best accuracy.
On a different dataset, \cite{attal2015physical} compared four supervised classifiers, i.e., KNN, SVM, Gaussian Mixture Models (GMM) and RF, as well as
several unsupervised models, showing that KNN achieve the best results among supervised solutions.
The authors of~\cite{bayat2014study} used a low-pass filter pre-processing followed by 5 classifiers, also investigating the combination of multiple classifiers. Finally, \cite{bianchi2019iot} proposed an innovative HAR system based on deep learning, using a Convolutional Neural Network (CNN) to recognize 9 different activities.

All aforementioned examples target high-end edge devices such as smartphones, proposing models with relatively large memory footprints and computational complexities. As a relevant example of a commercial HAR system for a more ``extreme'' edge, STMicroelectronics manufactured a system-in-package featuring a 3D digital accelerometer and gyroscope named LSM6DSOX~\cite{stsensor}. This system integrates a digital Machine Learning Core (MLC) which enables the on-chip identification of human activities such as running, walking and driving, based on the data patterns coming from the sensors. Specifically, HAR is realized with a feature extraction stage followed by a RF-based classifier, selected because of its HW-friendly operations.
More recently, DL approaches have been also proposed, obtaining state-of-the-art accuracies on several HAR datasets~\cite{hammerla2016deep,ordonez2016deep}. In order to obtain such results however, deep networks with parameters in the order of millions or billions have to be used. These models require far more computational power and memory than those available on low-power devices.

\subsection{IoT end-nodes and RISC-V}

IoT end-nodes are commonly based on general purpose CPUs or microcontrollers (MCUs) typically with a RISC instruction set, due to their high programmability, low power consumption and low cost~\cite{pulpNN}.
Customized hardware accelerators are rarely employed, since despite their exceptional performance and energy-efficiency, high manufacturing costs and limited flexibility make them convenient only for high-volume/high-price devices (e.g. smartphones), and not for lower-end systems (e.g. smartwatches, wearables, etc). Among general purpose solutions, the open-source RISC-V Instruction Set Architecture (ISA) is becoming widely popular in the IoT world. Several universities and companies have developed RISC-V-based cores with specific extensions targeting ML and DL at the edge~\cite{sifive,quentin}. 
A relevant example is the PULP family of processors~\cite{pulp,gap8}, on which our proposed BNN library is benchmarked in Section~\ref{sec:results}. Besides including features such as hardware loops, load/store with pointer increments and SIMD operations,
PULP processors are cacheless, leaving the full control of the memory hierarchy to the programmer. Despite increasing the software complexity, this feature, as well as the previous ones, yield significant speed-up and energy-efficiency benefits for the memory- and compute-intensive kernels involved in ML and DL inference.
Other RISC-V designers have also proposed similar features to enable deep learning at the edge~\cite{sifive, avispado}.

\subsection{Binary Neural Networks}

Quantization is known to be one of the most effective ways to reduce the inference complexity of DL models, which consists in reducing the precision of the data format used to represent model weights and layers inputs/outputs (so-called activations)~\cite{Hubara2017,Gupta2015,Jacob2018,JahierPagliari2018a}. The most common form of quantization uses 8-bit integer formats for both weights and activations, yielding relevant complexity reduction with limited accuracy drops with respect to floating point models~\cite{Hubara2017,JahierPagliari2018a}. Nonetheless, the memory footprint and total number of inference operations of 8-bit quantized models are often still excessive for extreme edge devices~\cite{hubara,xnornet}. 

Binary Neural Networks (BNNs) try to achieve further complexity reductions by taking the quantization concept to the extreme, reducing both weights and activations to 1-bit precision~\cite{hubara}. The most hardware-friendly form of binarization allows weights and activations to assume numerical values in $\{-1, 1\}$, which are then encoded in binary as logic 0 and 1, respectively. Thus, binarization is obtained simply with the $sign()$ function.
Binarizing both weights and activations with this format impacts the memory footprint of the model significantly, lowering it by 32x with respect to a sigle-precision floating point implementation. Similarly, the size of intermediate activation buffers is also reduced.

Furthermore, binarization also
completely eliminates Multiply and Accumulate (MAC) operations from the main computational kernels of BNNs (e.g., fully connected and convolutional layers) in favour of bit-wise operations. In fact, BNN multiplications can be implemented as bitwise XNORs, while accumulations reduce to counting the number of 1s (\textit{Popcount}). Precisely, a dot product between two binary vectors of weights ($\mathbf{w}$) and activations ($\mathbf{x}$) is obtained with the following equation:
\begin{equation}\label{eq:dotp_bin}
    y = 2 \cdot \mathlarger{\mathlarger{P}}_{i=0}^N w_i \odot x_i - N
\end{equation}
where $\odot$ and $\mathlarger{P}$ are the XNOR and popcount operators respectively, and $N$ is the vectors' length. Note that both the subtraction and the scalar multiplication are constant operations that in practice can be eliminated with no impact on the model's capacity.
The possibility of performing inference with these HW-friendly binary operations and of executing 32 ``multiplications'' in parallel even on a general purpose processor, by XNOR-ing together two 32-bit registers, enables significant speedups, even compared to 8-bit quantization~\cite{hubara,xnornet}.

\begin{figure*}[ht]
    \centering
    \includegraphics[width=.8\textwidth]{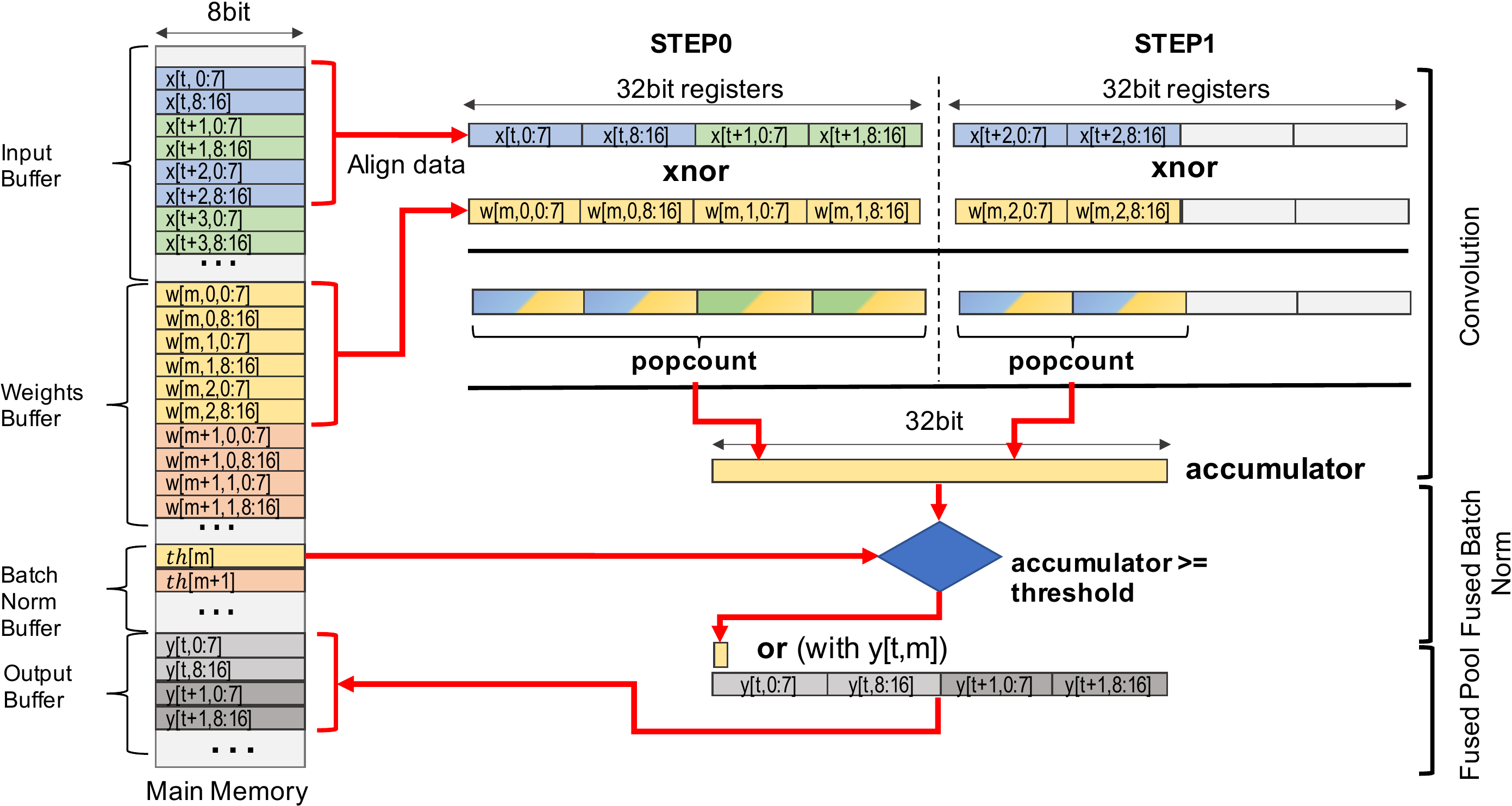}%
    \caption{High-level view of the operations in the proposed binarized convolutional layers for $C_{in} = C_{out} = 16$ and $K=3$.}\label{fig:conv_overview}%
    \vspace{-0.3cm}
\end{figure*}

BNNs are not a panacea, as they
often incur in sharp accuracy drops compared to float or 8-bit integer networks on difficult tasks, e.g., in computer vision~\cite{hubara}. Nonetheless, on simpler problems, they have been shown to yield excellent complexity vs. accuracy trade-offs~\cite{hubara, xnornet}.
For this reason, BNNs are promising candidates for an efficient yet accurate implementation of HAR on edge devices. However, to the best of our knowledge, no previous work has explored them for this task.

\section{Ultra-compact BNNs for HAR}\label{sec:method}
\subsection{Motivation}
In this paper, we pursue two related goals. First, we want to assess the effectiveness of BNNs for implementing HAR on edge devices, in comparison with state-of-the-art hardware-friendly ML models such as RFs. In particular, we target general purpose RISC-V processors, due to their increasing popularity in IoT. Since these devices often have limited memory and tight energy constraints~\cite{quentin}, in order to be competitive, our BNN implementation should be optimized to minimize the number of operations performed and the models' memory footprints. Therefore, our second goal is to introduce a novel BNN inference library that specifically targets ultra-compact networks.

In fact, 
a common design decision of previous BNN libraries for general purpose hardware is constraining the number of channels (i.e., feature maps) in each layer to multiples of 32~\cite{pulpNN,bmxnet,zhang2019dabnn}.
This permits the storage of all channels relative to the same feature map element exactly in one or more 32-bit words, thus significantly simplifying the implementations of convolutions on 32-bit architectures.
While this constraint is reasonable for 2D BNNs used in computer vision, which have many channels~\cite{hubara,xnornet}, our results of Section~\ref{sec:results} show that, for HAR, a good accuracy can be obtained with 1D BNNs including as little as 2 or 4 channels.
Therefore, sticking to existing implementations would force us to use over-parametrized models, resulting in a needless waste of operations and memory. For example, a binary 1D convolutional layer with $C_{in}=8$ input channels and $C_{out}=4$ output channels, with a filter size of $K=7$ requires $7\cdot 8 \cdot 4 = 224$ bits of storage. Being limited to $C_{in} = C_{out} = 32$ causes the data size to increase to $7168$ bits, i.e., a 31x overhead.
To cope with this limitation, our library efficiently supports any power-of-2 as number of channels, with minor overheads when dealing with larger networks.
The C implementation of our library is released open source at~\cite{code}.

In the rest of this section, we describe in detail our implementation of ultra-compact binarized layers.

\subsection{Binarized Convolution}

An overview of the flow of operations in our implementation of binarized convolutional (Conv) layers is shown in Figure~\ref{fig:conv_overview}.
The library focuses on 1D convolutions for time-series processing, as needed for HAR. Thus, the basic convolutional layer equation, assuming a stride $S=1$ for simplicity, is the following:
\begin{equation}\label{eq:conv}
y_{t,m} = f\left( \sum_{k=0}^K \sum_{c=0}^{C_{in}} w_{m,k,c} \cdot x_{t+k,c}\right)
\end{equation}
where $x$, $y$ and $w$ are input activations, output activations and filter weights respectively, $K$ the convolution filter size, $C_{in}$ the number of input channels and $f()$ a non-linear activation function, typically a ReLU. The computation is repeated for all output time-steps $t \in [0:T-1]$ and for all output channels $m \in [0:C_{out}-1]$.
If $x$ and $w$ are both binary, given (\ref{eq:dotp_bin}), the equation becomes:
\begin{equation}\label{eq:conv_bin}
y_{t,m} = sign\left(\mathlarger{\mathlarger{P}}_{k=0}^K \mathlarger{\mathlarger{P}}_{c=0}^{C_{in}} w_{m,k,c} \odot x_{t+k,c}\right)
\end{equation}
Notice that the (non-linear) $sign()$ function used to re-binarize the output also takes on the role of activation function~\cite{hubara}.

Equation (\ref{eq:conv_bin}) applies to all Conv layers, except the first layer of the network, i.e., the one that directly processes inertial sensor samples in case of HAR. Indeed, in accordance with several previous works~\cite{hubara,xnornet, zhang2019dabnn}, we found that binarizing the input directly yielded too low accuracy and divergent training. Therefore, 
our library also includes an implementation of mixed-binary-integer convolution, processing 8-bit quantized elements for $x$ (not detailed here for sake of space).
Moreover, notice that (\ref{eq:conv_bin}) does not include an additional bias term as the latter can be merged with batch normalization. (see Section~\ref{sec:bn_pool}).

\subsubsection{\textbf{Data Layout}}
To the best of our knowledge, all existing libraries implementing BNN inference on general purpose processors and MCUs focus on 2D convolutions~\cite{pulpNN,bmxnet,zhang2019dabnn}, whereas HAR requires 1D networks. Thus, a first important design choice of our implementation concerns the data layout in memory for 1D data, and in particular for the $T\times C_{in}$ activation tensors $X = [x_{t,c}]$.

In order to maximize performance, we store these data in \textit{time-major order}, i.e. all channels relative to the same time-step are stored contiguously, as shown in the \textit{Input Buffer} of Figure~\ref{fig:conv_overview}. Choosing this layout permits accessing \textit{all} inputs required to produce a given convolution output, i.e., the $K \times C_{in}$ bits relative to $K$ consecutive time-steps, by loading few consecutive words from memory. In turn, this allows us to exploit hardware features available in many RISC-V cores, such as loads with automatic post-increment. For example, with $C_{in} = 16$, $K=3$, all inputs required for a single convolution are stored as 48 consecutive bits, and can be loaded in two 32-bit registers (as shown in the top-right of Figure \ref{fig:conv_comparison}).

\subsubsection{\textbf{Loop Ordering}} The computation expressed by (\ref{eq:conv_bin}) has to be repeated for all output channels and time-steps, i.e., it has to be inserted within two nested loops. The ordering of these two loops is another important design choice, which determines the data reuse pattern, i.e., whether input activations or weights are kept fixed, while iterating over the other vector.
After testing both approaches, we found that keeping activations fixed yields the best performance on our target HW. Thus, we process all output channels relative to the same step sequentially before moving to the next time-step.  
The main advantage of this approach is that, since (binary) output activations relative to multiple channels of the same time-step are likely to be loaded in the same 32-bit register (see the bottom-right of Figure~\ref{fig:conv_overview}), output registers reuse is maximized. A pseudo-code of the overall layer processing is shown in Algorithm~\ref{alg:order}, where the inner loop operation corresponds to (\ref{eq:conv_bin}).

\begin{algorithm}[ht]
\SetAlgoLined
\For{$t \in [0:T-1]$}{
 \For{$m \in [0:C_{out}-1]$}{
   $y_{t,m} = \mathrm{Conv}(w_m, x_{t:t+K-1})$  // Eq. (3)
 }
}
\caption{Loop ordering in binarized Conv layers.}\label{alg:order}
\end{algorithm}

\subsubsection{\textbf{Data Alignment and Leftovers}}\label{sec:align}
The main difference between our library and previous approaches for 2D networks, which deal only with multiples of 32-channels, lies in the fact that convolution inputs may occupy a non-integer number of words. This is shown in Figure~\ref{fig:conv_overview}, where the inputs of a convolution occupy 48bit, i.e., 1 and 1/2 words for a typical 32-bit IoT processor. With the goal of minimizing operations and memory, padding inputs and weights buffer with zeros is not an option.

\begin{figure}[ht]
    \centering
    \includegraphics[width=0.7\linewidth]{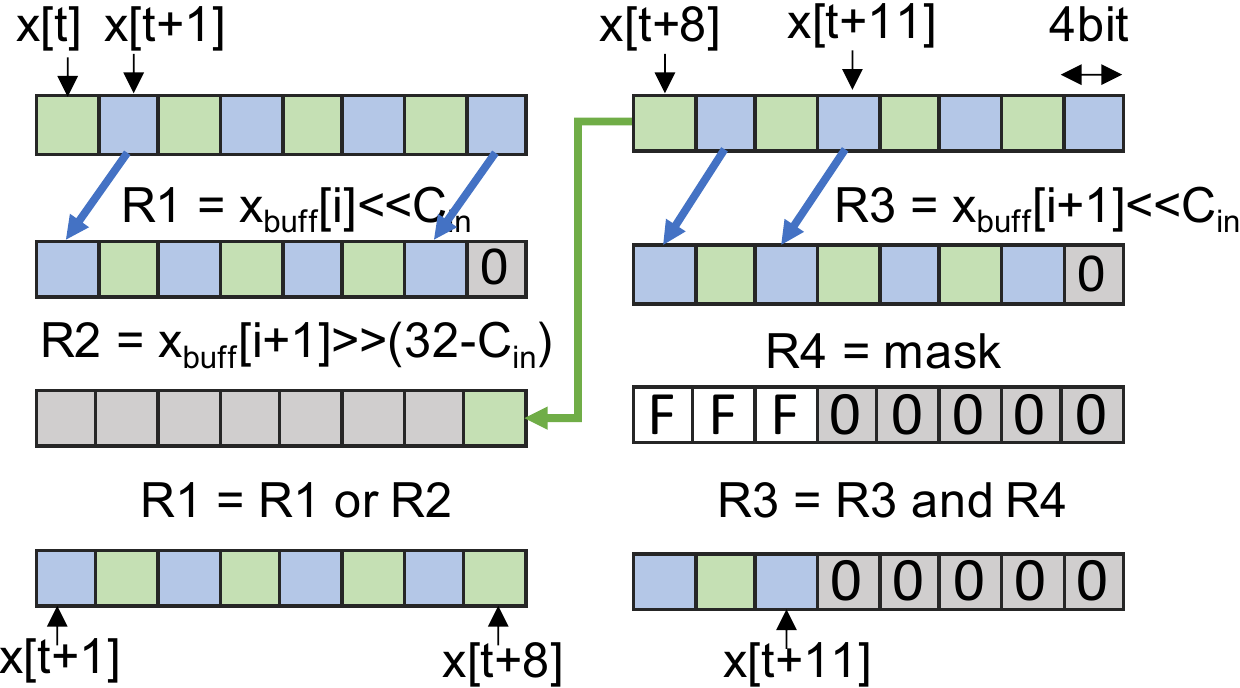}
    \caption{An example of input alignment for $C_{in} = 4$, $K = 11$.}
    \label{fig:alignment}
\end{figure}

It becomes then necessary to carefully handle data alignment and partial ``leftovers'', while maintaining as much as possible the parallelization benefits of binarization.
An example of how we handle the alignment of inputs is shown in Figure~\ref{fig:alignment} for $K=11$ and $C_{in} = 4$    . The figure refers to the moment when, after having produced the outputs relative to all channels of timestep $t$, the layer has to start processing timestep $t+1$, i.e., the end of an iteration of the outer loop in Algorithm~\ref{alg:order}. Thus, since $K=11$, the input samples to be considered change from $x[t:t+10]$ to $x[t+1:t+11]$. Given the chosen data layout, to correctly align the inputs against the filter weights, all input samples have to be shifted left by $C_{in}$ positions. Further, for each whole word involved in the convolution, the most significant $C_{in}$ bits of the next word have to be shifted in, as shown in the left of Figure~\ref{fig:alignment}. Finally, the leftover partial word has to be filtered with a mask containing $(K \cdot C_{in}) \% 32$ leading 1s, to avoid that XNOR and popcount operations are applied also to inputs not involved in this convolution (right side of the figure). A similar alignment step, not reported for sake of space, is necessary also for filter weights, when switching from one output channel to the next.

To avoid destructing input data, aligned inputs and weights are stored in intermediate buffers (composed of 2 words in the example of Figure~\ref{fig:alignment}), when they cannot be entirely kept in the register file.

\subsubsection{\textbf{Partial Loop Unrolling}}\label{sec:unroll}
Our Conv implementation also uses partial loop unrolling, i.e., the processing of multiple time-steps and/or output channels in each inner loop iteration of Algorithm~\ref{alg:order}. As explained in previous work~\cite{pulpNN}, this improves the arithmetic intensity, that is, for a BNN, the number of XNOR/Popcount operations per memory load. However, an excessive unrolling would cause register spilling, yielding the opposite effect. After experimenting with different unroll patterns, we found that the best performance on our target hardware are achieved with a 2x2 kernel, i.e., processing 2 output channels relative to 2 consecutive time-steps in each inner loop iteration. 

\begin{figure}[ht]
    \centering
    \includegraphics[width=.8\linewidth]{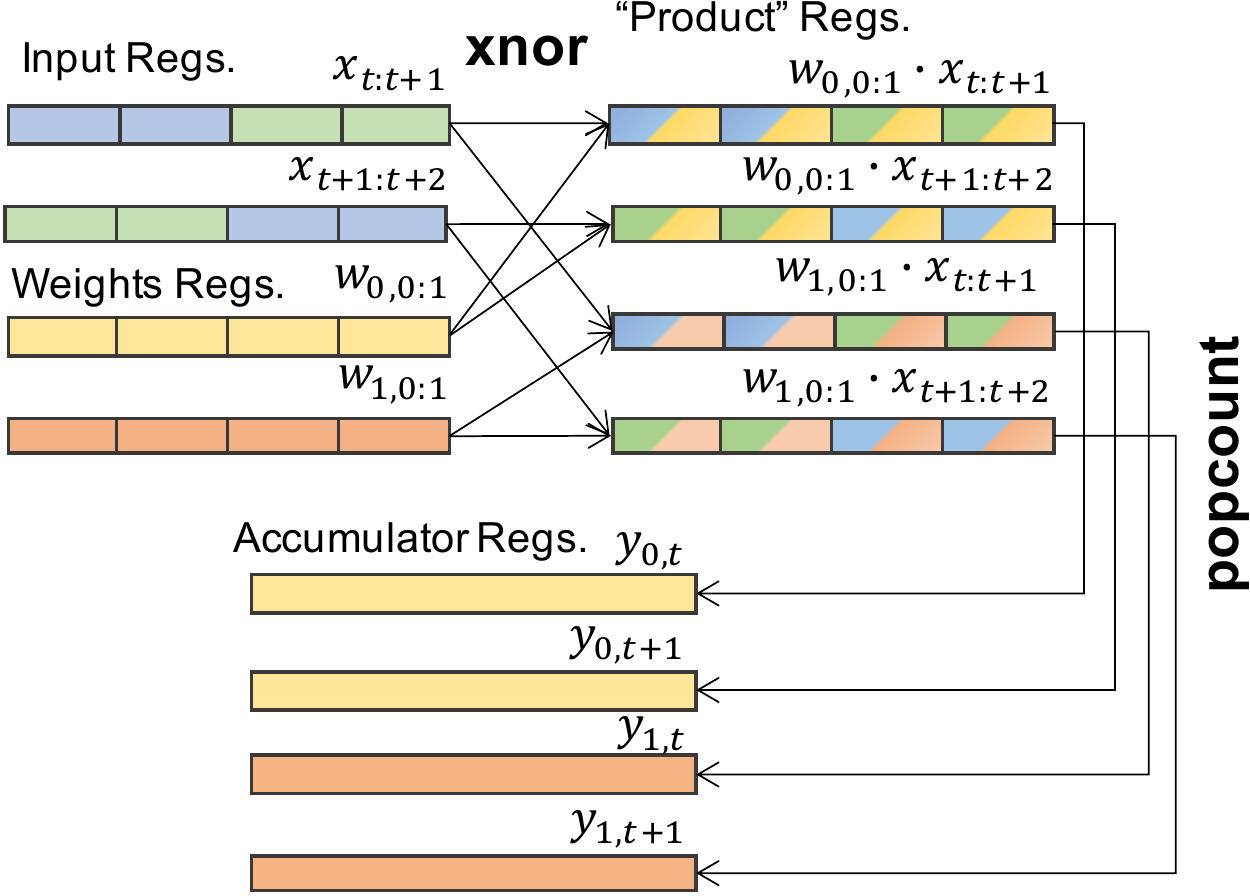}
    \caption{Partial loop unrolling with a 2x2 kernel.}
    \label{fig:unrolling}
\end{figure}

A schematic view of the 2x2 inner loop processing is shown in Figure~\ref{fig:unrolling}, for the first step of a convolution with $C_{in} = 16$ (further steps would be necessary if $K>2$). As shown, 12 32-bit registers are involved in the computation: 2 each for inputs and weights, 4 for the XNOR outputs (products) and finally, 4 accumulators storing the (pre-binarization) outputs relative to the two processed channels (0 and 1) and timesteps ($t$ and $t+1$).

\subsection{Fused batch normalization and pooling}\label{sec:bn_pool}

Batch Normalization (BatchNorm) is a common operation in modern neural networks. For a BNN, BachNorm is typically performed before re-binarizing outputs. In this case, as explained in~\cite{hubara}, the combined BatchNorm and binarization can be reduced to a simple comparison between the pre-binarization output relative to each channel and a threshold computed offline as follows:
\begin{equation}\label{eq:batchnorm}
  th_m = \frac{(\mu_m - \beta_m \cdot \frac{\sigma_m}{\gamma_m}) + K\cdot C_{in}}{2}     
\end{equation}
where $\mu_m$ and $\sigma_m$ are the mean and standard deviation values relative to the $m$-th channel, and $\beta_m$ and $\gamma_m$ are the trained BatchNorm parameters.

In our library, BatchNorm is therefore implemented as a \textit{fused} operation within convolutional and fully connected layers, as shown in the bottom-right of Figure~\ref{fig:conv_overview}. This minimizes the memory accesses by avoiding the storage of temporary (not-binarized) accumulator outputs in a large intermediate buffer.

Similarly , in order to reduce the latency and code size of our implementation, we also fuse max pooling (MaxPool) operations with convolutional and fully-connected layers. In BNNs, MaxPool reduces to a bitwise OR among the outputs relative to multiple consecutive timesteps, which can be easily implemented ``on the fly'' by means of a correct alignment on the output register (as shown in Figure~\ref{fig:conv_overview}).
Doing so reduces the read/write operations on the output buffer, as well as its size. Most importantly, it eliminates the need of a dedicated max pooling function, with a positive effect on code size.

\subsection{Binarized Fully-Connected Layer}

Our implementation of fully connected (FC) layers follows very similar principles to those described for convolutions, not repeated here for sake of space. In fact, a FC layer can be considered as a corner case of convolution with a single output timestep. The main difference is that, for FC layers, the partial loop unrolling described in Section~\ref{sec:unroll} and shown in Figure~\ref{fig:unrolling} is performed with a 2x1 kernel, since we don't have multiple ``timesteps'' to process.

Importantly, most of our BNNs for HAR include a \textit{single} FC layer at the output of the network, used to produce the final classification scores. For this layer, we skip the re-binarization of the outputs, saving the 32bit accumulator output directly to memory, since the argmax of the scores is used to determine the predicted class. This, in turn, affects the implementation of BatchNorm which, for the output layer, is not realized by a simple thresholding as in (\ref{eq:batchnorm}) but rather with a standard product + sum.

\section{Experimental Results}\label{sec:results}

\subsection{Setup}\label{sec:setup}
We perform experiments on two different HAR datasets, one public and one proprietary.
The former, \textit{UniMiB-SHAR}~\cite{micucci2017unimib} contains 11,711 records acquired with an Android smartphone, divided in 17 activity classes. Each record is a vector of 151x3 tri-axial accelerometer values. In the original paper, the authors benchmarked this dataset with multiple classifiers using either raw data or acceleration magnitude features, and obtained the best performance with a random forest (RF).
We then also experiment on an internal proprietary dataset, containing 667 records, each formed by multiple non-overlapping windows of 32 tri-axial accelerometer samples. In this case, the HAR target is simpler, and just consists in a binary distinction between ``walking'' and all other activities. 
On this dataset, which we refer to as \textit{Walk} in the following, the previous best results were also obtained applying a RF model, and using the following 7 features, computed on each of the 3 accelerometer axes over each 32 samples window: average, variance, energy, max, min, peak-to-peak and number of zero-crossings.

We deployed all models on the 32bit single-core RISC-V platform PULPissimo, based on the RI5CY core~\cite{pulp}. Inference clock cycles are estimated with the virtual platform GVSOC~\cite{gvsoc}, whereas energy results refer to the 22nm implementation of PULPissimo found in~\cite{quentin}, called Quentin.
BNNs are trained in PyTorch, whereas for RF baselines we use the scikit-learn Python package. All PULPissimo implementations are in C language.

\subsection{Comparison baseline}

To prove the effectiveness of BNNs for HAR, we compare them against RFs, which are widely considered state-of-the-art HW-friendly models for such task~\cite{stsensor,micucci2017unimib}. 
For a fair comparison, our RF implementation should also be optimized. Since we couldn't find an open source library for the same HW target, in this section we briefly describe our in-house RF implementation. The latter is roughly inspired by the one of OpenCV~\cite{opencv_library}, albeit stripped-down and optimized for low-memory devices. 
A RF is represented by multiple C arrays, used instead of lists (as in OpenCV) for better memory locality and simpler control flow. The RF processes an \texttt{INPUT} matrix containing either raw data or extracted features (depending on the experiment).

\begin{figure}[ht]
    \centering
    \begin{subfigure}{0.99\linewidth}
    \includegraphics[width=\linewidth]{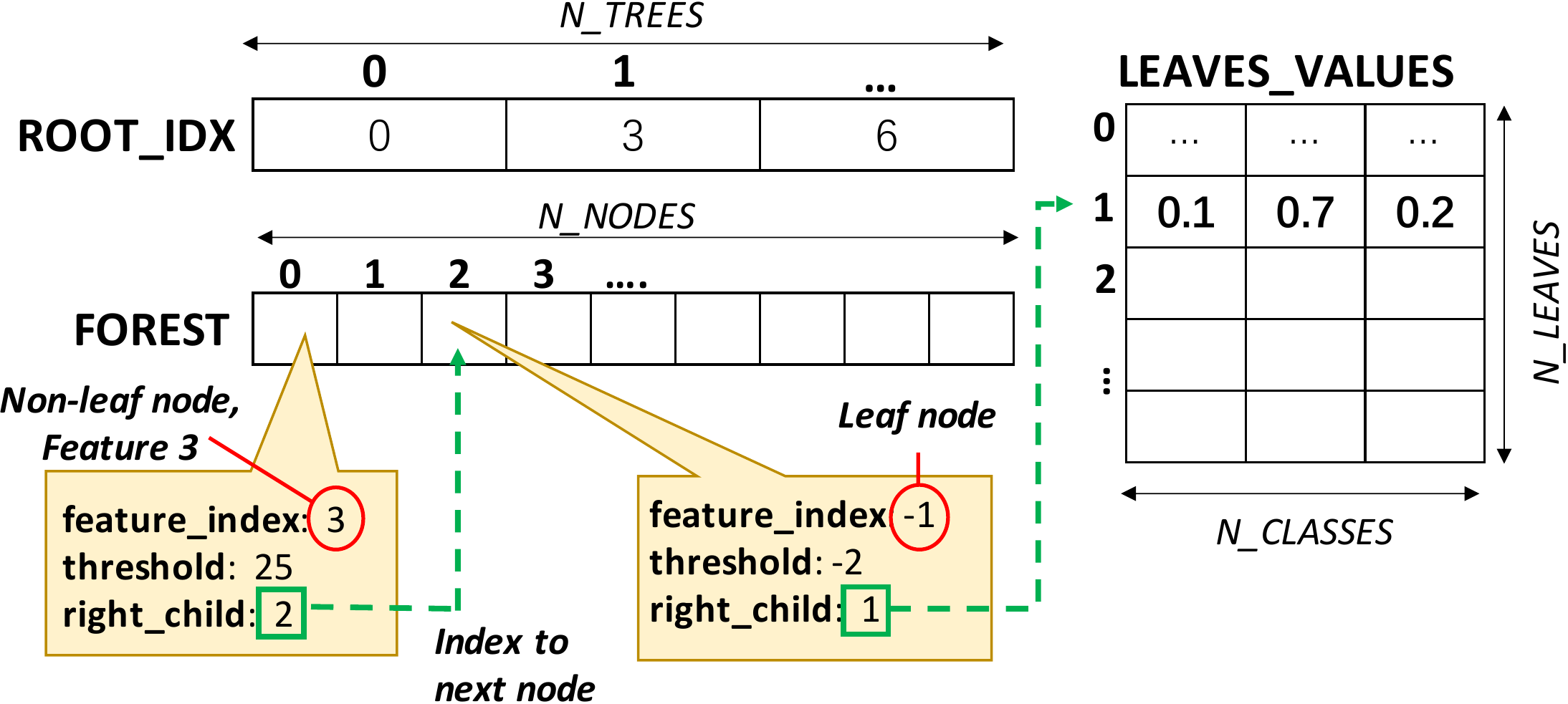}
    \subcaption{Array representation}\end{subfigure}
    \begin{subfigure}{0.7\linewidth}
    \includegraphics[width=\linewidth]{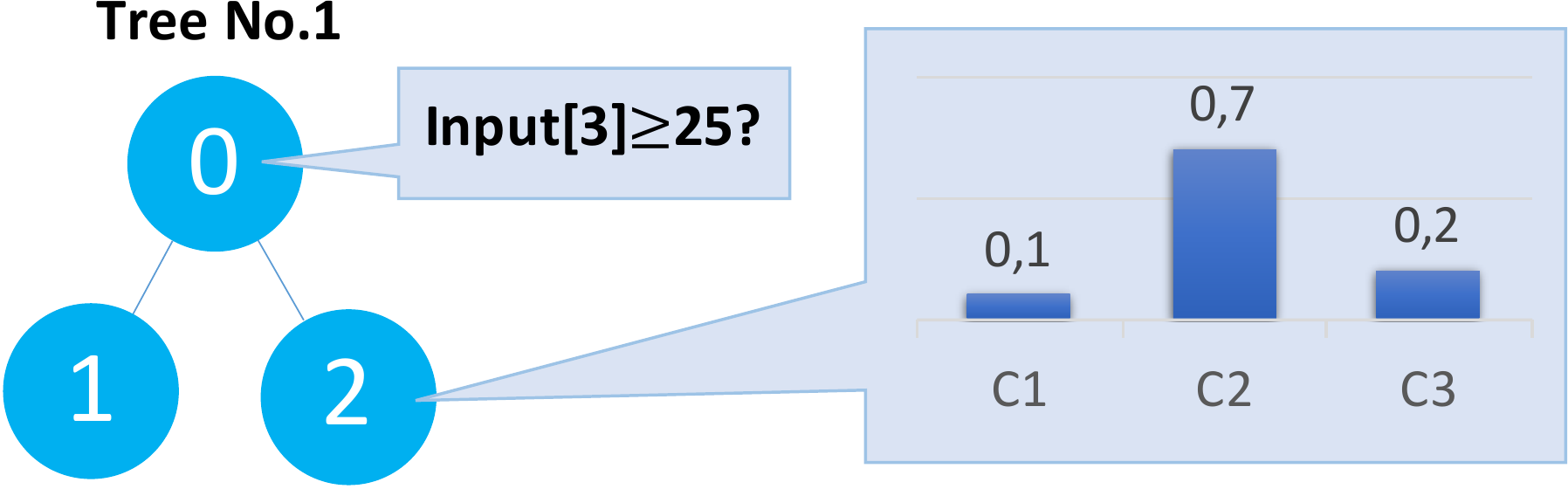}
    \subcaption{Corresponding (partial) RF}\end{subfigure}
    \caption{RF implementation used as comparison baseline.}\label{fig:rf}
    \vspace{-0.3cm}
\end{figure}

The main data structure representing the model is the \texttt{FOREST} array, which contains the description of all nodes in the RF. Its elements are C ``structs'' with three fields:
\begin{itemize}
    \item \texttt{feature\_index}: a column index in \texttt{INPUT} corresponding to the feature that should be compared against the threshold in the node. Leaf nodes are identified by a -1 in this field.
    \item \texttt{threshold}: the comparison threshold used in the node.
    \item \texttt{right\_child}: the index of the right child of the node in \texttt{FOREST}. To save memory, only the right child is stored, whereas the left child is implicitly equal to the following node in the array. For leaf nodes, \texttt{right\_child} is an index in the \texttt{LEAVES} array.
\end{itemize}
The \texttt{LEAVES} array stores the output prediction for all leaf nodes. Specifically, our implementation stores \textit{class probabilities} for all classes in the leaves, rather than a single class label, as this yields higher accuracy and also matches the scikit-learn implementation, making the conversion of models easier. Finally, the \texttt{ROOT} array stores the indexes of the roots of all trees in \texttt{FOREST}. An example of our implementation is shown in Figure~\ref{fig:rf}, highlighting the content of some \texttt{FOREST} elements relative to the first tree of the RF.

Inferences are initiated looping over all elements in \texttt{ROOT} and
visiting each tree.
The predicted probabilities of the different trees are accumulated and a final \textit{argmax} determines the class label.
Thresholds and probabilities are quantized to 8bit integer, since RI5CY does not have a FPU. Moreover, indexes in the different arrays also use 8 or 16bit, depending on the maximum RF size, to minimize the memory footprint.

\subsection{Comparison with existing BNN libraries}
In order to compare our library against a state-of-the-art BNN implementation for RISC-V, we adapted the open source binarized 2D convolutions of~\cite{pulpNN} to process 1D data, removing the so-called ``im2col'' phase, and modifying it to work on the single-core PULPissimo. Both libraries have been then compiled with identical flags.
Since~\cite{pulpNN} is limited to multiples of 32 channels, its weights and activations buffers have been zero-padded when processing layers with fewer channels.
\begin{figure}[ht]
\centering
\includegraphics[width=.6\linewidth]{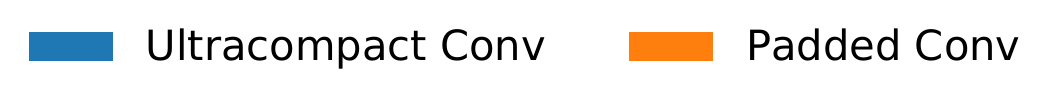}

\begin{subfigure}{\linewidth}
\includegraphics[width=.47\linewidth,trim=0cm 0cm 0cm 0cm,clip]{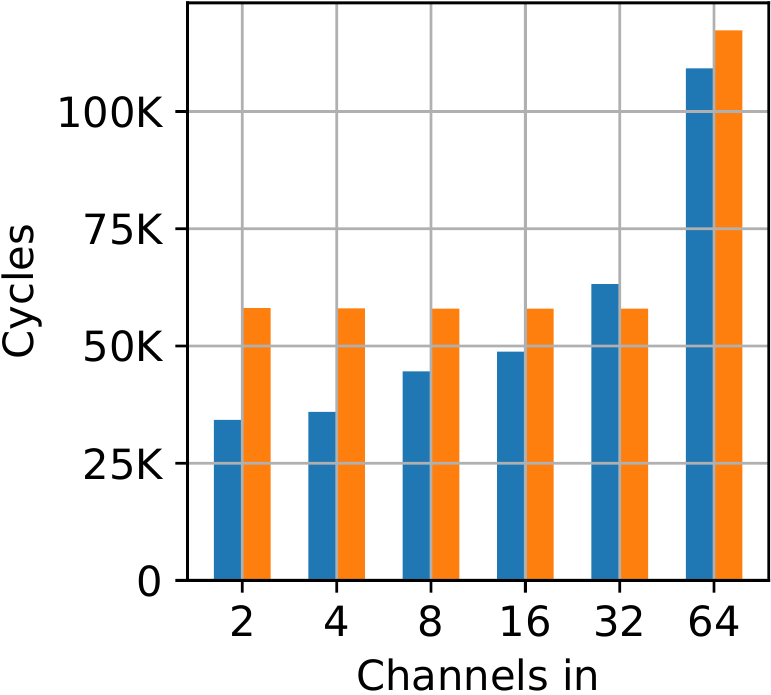}
\includegraphics[width=.47\linewidth,trim=0cm 0cm 0cm 0cm, clip]{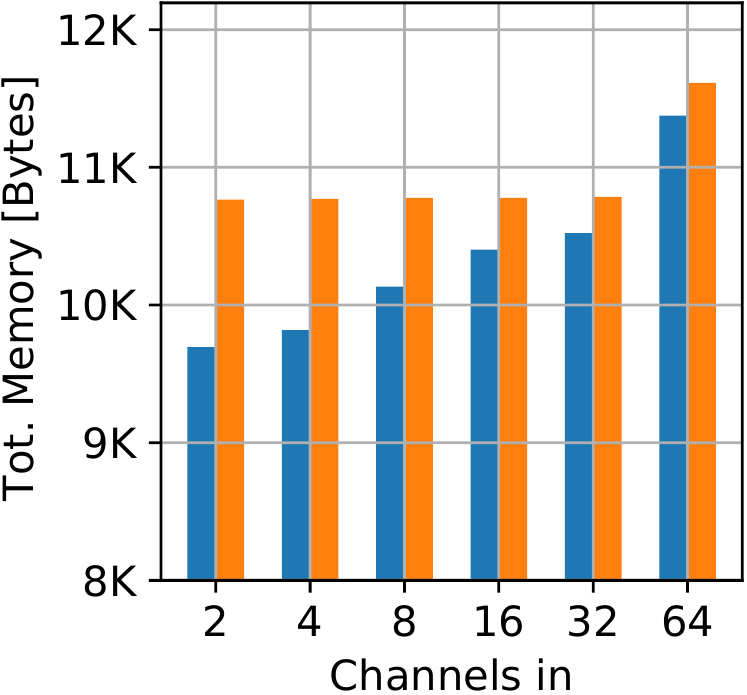}
\caption{$C_{out} = 8$}
\label{fig:cycles_conv_comparison}
\end{subfigure}

\begin{subfigure}{\linewidth}
\includegraphics[width=.47\linewidth,trim=0cm 0cm 0cm 0cm,clip]{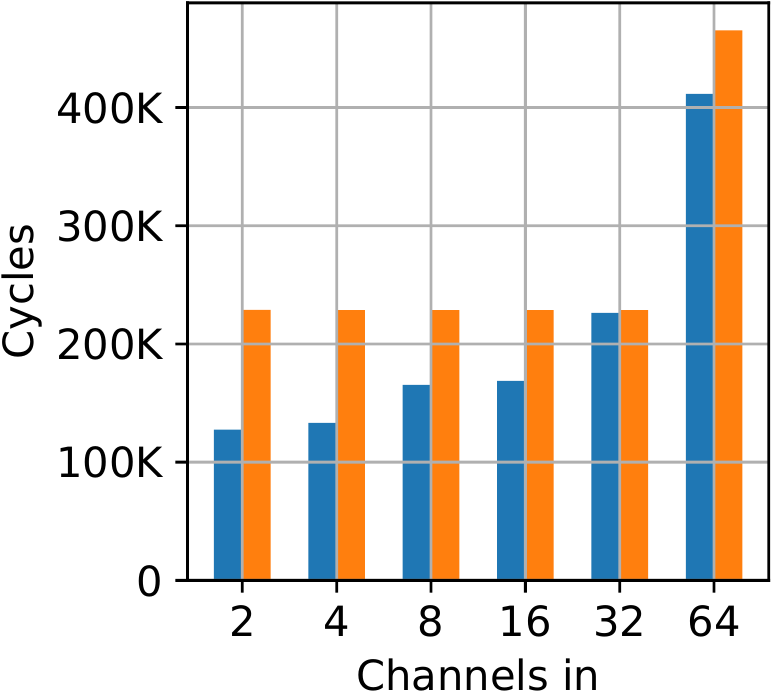}
\label{fig:cycles_conv_comparison8CH}
\includegraphics[width=.47\linewidth,trim=0cm 0cm 0cm 0cm, clip]{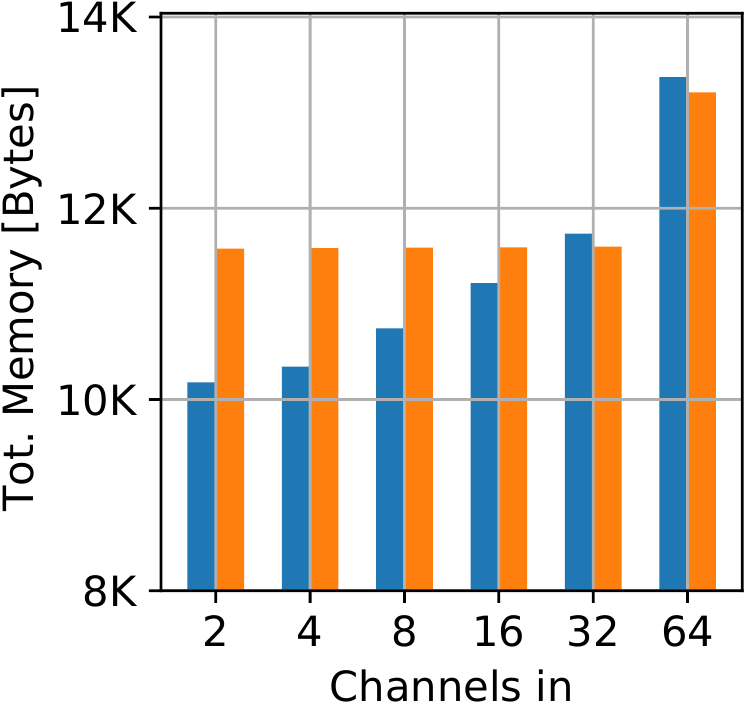}
\subcaption{$C_{out} = 32$}
\end{subfigure}
\caption{Comparison between our Ultracompact Convolution implementation and \cite{pulpNN}. Total clock cycles per inference and total memory occupation versus $C_{in}$, for two values of $C_{out}$, and averaged over multiple layers with, $K \in \{3,5,7\}$, and $T \in \{32, 64, 128, 256\}$.}
\label{fig:conv_comparison}
\end{figure}

Figure~\ref{fig:conv_comparison} shows the results of this comparison in terms of inference clock cycles and total memory occupation (code+data). The four graphs report the cycles and memory for different values of $C_{in}$. The two on top refer to $C_{out} = 8$, whereas the two on the bottom to $C_{out} = 32$. Cycles and memory are then averaged over different values of $K$ and $T$.

As expected, due to the useless operations performed with padded activations and weights, our library outperforms~\cite{pulpNN}, reducing the cycles by up to 41/44\% for $C_{in} = 2$ and $C_{out}=8/32$. In contrast, for $C_{in} \ge 32$, padding isn't required. However, our library is still  slightly faster than~\cite{pulpNN} in most cases, mostly due to a better selection of the partial unrolling size in our implementation.

\begin{figure*}[ht]
\includegraphics[width=.75\textwidth]{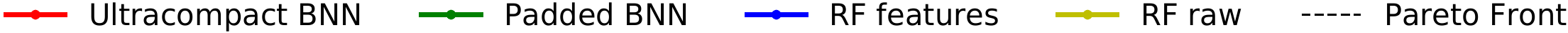}
\vspace{.5cm}

\begin{subfigure}{0.32\linewidth}%
\includegraphics[width=\linewidth, trim=0cm 0cm 0cm 0cm, clip]{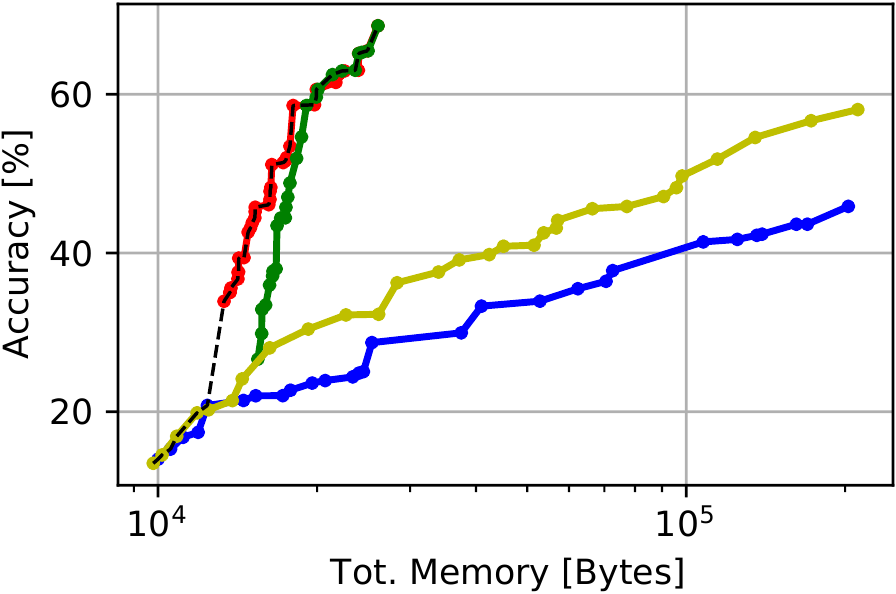} %
\end{subfigure}%
\hspace{0.5cm}
\begin{subfigure}{0.32\linewidth}%
\includegraphics[width=\linewidth, trim=0cm 0cm 0cm 0.0cm clip]{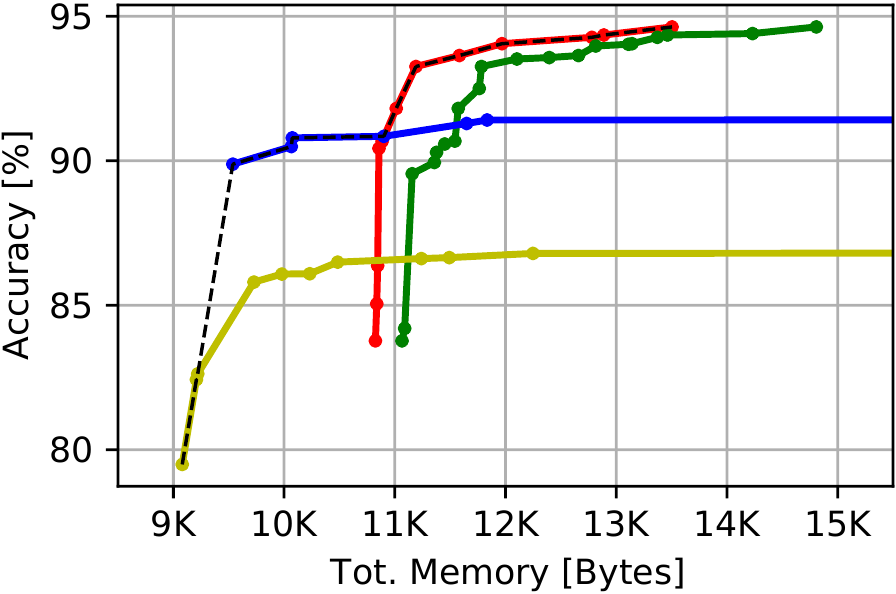} %
\end{subfigure}

\begin{subfigure}{0.32\linewidth}
\includegraphics[width=\linewidth, trim=0cm 0cm 0cm 0cm, clip]{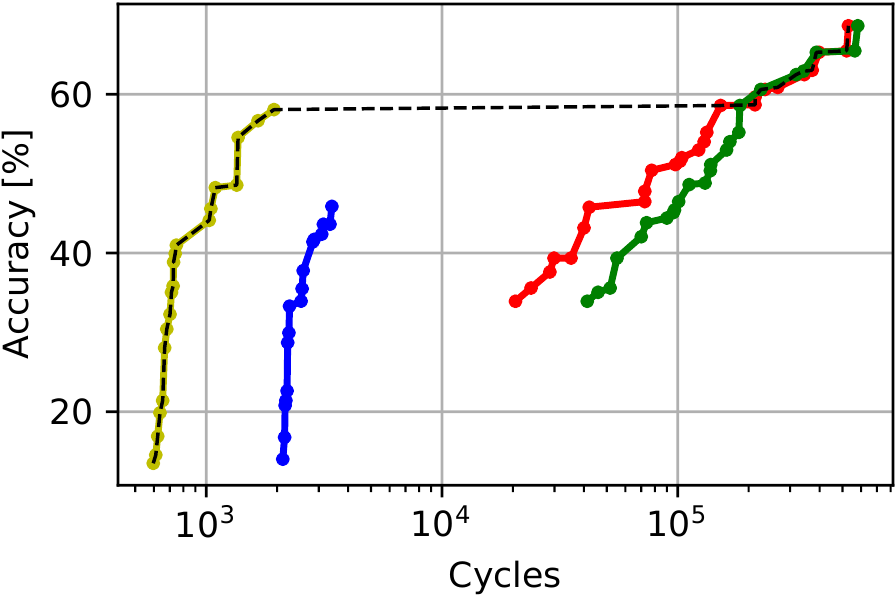}
\caption{UniMiB-SHAR}
\label{fig:unimib_mem_comparison}
\end{subfigure}
\hspace{0.5cm}
\begin{subfigure}{0.32\linewidth}
\includegraphics[width=\linewidth, trim=0cm 0cm 0cm 0.0cm, clip]{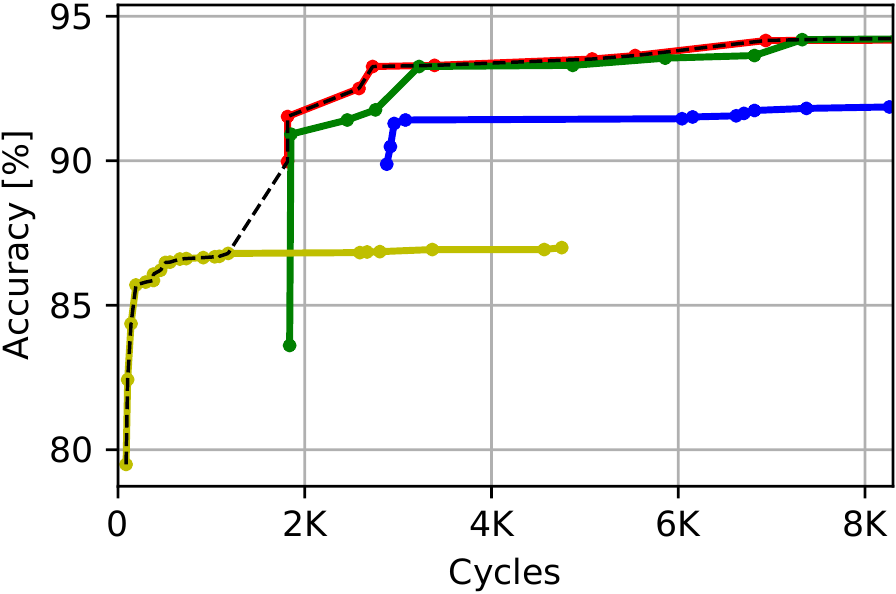}
\caption{Walk}
\label{fig:st_mem_comparison}
\end{subfigure}
\caption{Accuracy versus total inference clock cycles and total memory occupation of different RFs and BNNs implementation. Each point represents a different configuration of hyperparameters.}
\label{fig:unimib_and_st_comparison}
\end{figure*}

Similarly, we also reduce the memory occupation with respect to~\cite{pulpNN} for all convolutions with $C_{in} < 32$. The total memory saving reaches 10/12\% for $C_{in}=2$ and $C_{out}=8/32$. This smaller reduction is due to the code memory being the dominant contributor with respect to weights and activations buffers; considering only data memory the saving is $>70\%$ in both cases. For $C_{in} \geq 32$, we obtain a slight memory overhead due to the additional alignment buffers, not needed in~\cite{pulpNN}. However, our library explicitly targets ultra-compact BNNs, therefore this small overhead on larger networks is acceptable.

\subsection{Comparison with the State-of-the-art}

Figure~\ref{fig:unimib_and_st_comparison} shows a comparison between RF and BNN-based classifiers on the two target HAR datasets. Results are reported as Pareto frontiers in the accuracy versus cycles and accuracy versus total memory planes, and refer to end-to-end inferences, i.e., including both feature extraction and classification.
For each classifier, different points correspond to Pareto-optimal sets of hyperparameters, determined with an extensive grid search~\footnote{Notice that the points in the cycles and memory Pareto curves are not necessarily the same, as one configuration may be Pareto-optimal in terms of cycles and not memory or vice versa.}. For RFs, we varied the number of trees and their maximum depth, whereas for BNNs we explored the networks' depth (number of layers) as well as the type and parameters of each layer ($C_{in}$, $K$, etc).
For both datasets, we compared our library (\textit{Ultracompact BNN}) with the state-of-the-art BNN implementation of~\cite{pulpNN} (\textit{Padded BNN}) as well as with two RF variants, one applied directly to raw data (\textit{RF raw}) and the other to the features detailed in Section~\ref{sec:setup} (\textit{RF features}). For the latter, cycles and memory results also include feature extraction. The black dashed line in the plots highlights the global Pareto-frontier considering all classifiers together. Notice that the two graphs relative to the UniMiB-SHAR dataset are semi-logarithmic.

On the \textit{UniMiB-SHAR} dataset, RF models achieve up to 60\% accuracy with a very small number of cycles, using a small number of trees with high depth. However, this depth comes at the cost of an exponential memory increase (up to $> 200$ kB). Interestingly, \textit{RF raw} outperforms \textit{RF features}, probably because the simple magnitude features proposed in~\cite{micucci2017unimib} are not very informative for the model.
On the other hand, BNNs have a higher initial baseline memory ($\approx 12$ kB) due to the larger code size, but then sharply improve the accuracy with a small increase in memory occupation. In terms of cycles, RFs clearly outperform BNNs for accuracy values $< 60\%$, requiring around 2.5k cycles even for large models. However, we could not find a RF-based solution able to produce higher accuracy, without exceeding the entire 520kB of memory of the target HW. In contrast, BNNs reach up to $\approx$70\% accuracy (10\% improvement over RFs) with $<$30kB of total memory and $\approx$ 500k cycles. At 59\% accuracy, i.e. the maximum reached by RF raw, the latter requires 211kB of memory while our BNN library only occupies 18kB, resulting in a 91\% memory reduction.
Our BNN library also outperforms the implementation in~\cite{pulpNN}, reducing the total memory and cycles by up to 7\% and 21\% respectively, for the same accuracy. Crucially, some of the BNN configurations are on the global Pareto-front (black dashed line) in both planes.

\begin{table*}[ht]
\centering
\caption{Detailed deployment results of the proposed Ultracompact BNNs.}\label{table:results}
\footnotesize
\begin{tabularx}{\linewidth}{ c|c|c|c|c|c|X}

\textbf{Dataset} & \textbf{Config.} &  \textbf{Accuracy} & \textbf{Tot. Memory [kB]} & \textbf{Tot. Energy [$\mu J$]} & \textbf{Latency [$\mu s$]} & \textbf{Architecture} \\ \hline
 \multirow{4}{*}{Walk} & Min & 0.837 & 10.82 & 0.036 & 9.36 & Conv(2,7), Conv(2,15), Pool(4,4), FC\\\cline{2-7}
  & Max - 10\% & 0.85 & 10.83 & 0.035 & 9.22 & Conv(2,7), Conv(2,15), Pool(2,2), FC \\\cline{2-7}
  & Max - 5\% & 0.9 & 10.85 & 0.04 & 10.4 & Conv(2,7), Conv(2,7), Pool(2,2), FC \\\cline{2-7} & Max & 0.946 & 13.5 & 0.45 & 118.8 & Conv(8,7), Conv(32,15), Pool(4,4), FC \\  \hline\hline %
 \multirow{4}{*}{UniMiB-SHAR} & Min & 0.33 & 13.32 & 0.38 & 99.94 & Conv(4,7), Conv(4,7), Pool(4,4), Conv(4,7), Pool(4,4), FC\\\cline{2-7}
  & Max - 10\% & 0.586 & 19.7 & 3.95 & 1037.22 & Conv(8,15),Conv(32,7),Pool(4,4), Conv(32,7), Pool(4,4), FC \\\cline{2-7}
  & Max - 5\% & 0.651 & 23.98 & 9.08 & 2385.31 & Conv(32,7),Conv(32,15),Pool(4,4), Conv(32,15), Pool(4,4), FC \\\cline{2-7}
  & Max & 0.68 & 26.07 & 9.84 & 2583 &  Conv(32,15),Conv(32,15),Pool(4,4), Conv(32,15), Pool(4,4), FC \\ \hline
\end{tabularx}
\end{table*}

On the simpler \textit{Walk} dataset, all models require less cycles and memory. When comparing RFs and BNNs, the main difference is due to the more informative features extracted for this dataset. Thanks to this more elaborated feature extraction step, \textit{RF features} models obtain higher accuracies compared to \textit{RF raw} ones; however, feature extraction also increases the cycles of former significantly, making them completely Pareto-dominated by BNNs. As for \textit{UniMiB-SHAR}, our BNN library obtains the highest accuracy overall (up to $\approx 95\%$), outperforming a padded implementation. Furthermore, BNN solutions appear again on both Pareto frontiers, demonstrating the effectiveness of these models for HAR.

\subsection{Detailed deployment results}

The wide ranges of clock cycles and memory footprints achieved by our ultra-compact BNNs for HAR translate into very different accuracy versus latency and accuracy versus energy trade-offs when the networks are deployed on an edge processor. This is shown in Table~\ref{table:results}, which details the metrics of the two extreme BNN Pareto points of Figure~\ref{fig:unimib_and_st_comparison} for both datasets, i.e., those with the minimum (Min) and maximum (Max) accuracy. Furthermore, the smallest models able to achieve a $<$5\% and $<$10\% accuracy drop compared to ``Max'' are also shown. For each model, the table reports the accuracy in $[0:1]$, the total memory footprint (code + data), and the total latency and energy consumption per inference.
Latency and energy results have been derived from the power values in~\cite{quentin} at 205MHz, assuming that the processor is power gated after an inference is completed, while the cycles have been estimated simulating the hardware with GVSOC~\cite{gvsoc}.

Finally, the table also shows the hyper-parameters configurations corresponding to each deployed BNN, as a comma-separated sequence of layers. In particular, Conv($C_{out}$,$K$) denotes a convolutional layer with $C_{out}$ output channels and kernel size equal to $K$, Pool($K$,$S$) denotes a MaxPool layer with stride to $S$ and kernel size $K$, and FC denotes a fully connected layer with a number of output neurons equal to the number of classes of the dataset. Notice that, as explained in Section~\ref{sec:method}, the first convolutional layer of each network is actually processing 8bit integer inputs. 

The small latency values ($<3ms$ even for the largest models) demonstrate that all deployed BNNs are suitable for a real-time implementation of HAR. In fact, the latency constraint for real-time classification is given by the input 
sampling period, which for UnMiB-SHAR and Walk corresponds to 20ms and 40ms, respectively. This result is achieved while consuming a very limited amount of energy ($<10\mu$J) per inference.

Despite the different complexity of the two datasets, in both cases different BNN configurations can span a significant range of accuracies, 11\% and 35\% for Walk and UniMib-SHAR respectively, with a corresponding increase of 15x and 26x between the minimum and maximum energy per inference.
With respect to RFs, the energy benefits at iso-accuracy reach up to 70\% for the \textit{Walk} dataset at 92\% accuracy, i.e., the point where the total cycles difference is maximal (see Figure~\ref{fig:conv_comparison}).
Globally, these results show that BNNs are able to obtain both acceptable accuracies with a very low energy consumption, but also highly accurate results at a still moderate energy cost, offering a wide spectrum of solutions to select based on application needs.

\section{Conclusions}
In this paper, we applied BNNs to HAR for the first time. 
We described in detail an optimized BNN implementation that targets ultra-compact models (i.e., with fewer than 32 channels per layer), to be deployed on edge devices. In comparison with state-of-the-art hardware-friendly models (RFs) we not only obtain higher accuracy, but also save up to 70\% energy (or 91\% memory) at iso-accuracy, depending on the target dataset. Moreover, we showed that BNN-based solutions can cover a wide range of accuracy versus energy and accuracy versus latency trade-offs, among which designers can select based on their needs.

\end{document}